\documentclass{article}
\usepackage{spconf,amsmath,graphicx, cite, svg, tabularx}
\usepackage{indentfirst}
\graphicspath{ {./Image/} }

\title{Gated Self-supervised Learning For Improving Supervised Learning}
\name{ Erland Hilman Fuadi, Aristo Renaldo Ruslim, Putu Wahyu Kusuma Wardhana, Novanto Yudistira
}
\address{Informatics Department, Faculty of Computer Science, Brawijaya University}

\begin{document}
%
\maketitle
\begin{abstract}


In past research on self-supervised learning for image classification, the use of rotation as an augmentation has been common. However, relying solely on rotation as a self-supervised transformation can limit the ability of the model to learn rich features from the data. In this paper, we propose a novel approach to self-supervised learning for image classification using several localizable augmentations with the combination of the gating method. Our approach uses flip and shuffle channel augmentations in addition to the rotation, allowing the model to learn rich features from the data. Furthermore, the gated mixture network is used to weigh the effects of each self-supervised learning on the loss function, allowing the model to focus on the most relevant transformations for classification.
\end{abstract}
\section{Introduction}
\label{sec:intro}

Recently, self-supervised learning \cite{Chen2020, He2020} has shown significant results in feature learning. Self-supervised learning is a promising approach to fixing the fundamental problem in training machine learning models where labeled data is scarce or expensive. For self-supervised learning, one of the most important techniques is data augmentation. In the case of image classification, self-supervised learning involves giving the model a set of unlabeled images and asking it to predict some properties of the images \cite{moon2022tailoring, Gidaris2018}. By predicting the images, the model can learn valuable features for classification. 

Using data augmentation techniques in image recognition enhances the model's ability to generalize by teaching insensitive features to spatial changes. This can be achieved by applying various modifications, such as geometric transformations (cropping, flipping, rotating) and photometric adjustments (brightness, contrast, and color). Recent methods have demonstrated the best performance, balancing complexity with accuracy and robustness. In addition, various methods for feature learning have specific to certain locations, which can also be used to transfer knowledge to tasks related to localization, like object detection or image labeling. These features allow the model to understand what and where to focus on making correct predictions.

The rotation has been attempted to be used for these augmentations, but relying solely on rotation as a self-supervised transformation can limit the ability of the model to learn rich features from the data. In this paper, we propose to use additional localizable augmentations, such as flip and shuffle channels, to provide a more diverse set of self-supervised signals. These augmentations allow the model to learn more complex features relevant to classification.

To ensure that the model focuses on the most relevant augmentations for classification, we use a Mixture of Expert (MoE), which gates the effects of each augmentation on the loss. This allows the model to dynamically adjust the importance of each augmentation, allowing it to focus on the most useful transformations for classification.

By utilizing these additional augmentations and the MoE method, our proposed approach can improve performance on image classification benchmarks, allowing the model to learn more complex features relevant to classification. Furthermore, our approach can improve the performance of other computer vision tasks, such as object detection and image annotation, as the learned features are transferable and localizable. Additionally, our approach can be used to improve the performance of models trained with scarce or expensive labeled data. Overall, our proposed method of using additional localizable augmentations and the MoE method to adjust the weighting of their importance can effectively improve self-supervised representation learning and computer vision tasks.

\section{Related Work}
\label{sec:format}

\textbf{Self-Supervised Learning:} Self-supervised learning has gained significant interest in recent years. It aims to learn general characteristics by solving tasks explicitly created for this purpose, known as pretext tasks. Depending on the number of examples used for these tasks, self-supervised learning can be divided into relation-based and transformation-based. Relation-based approaches focus on increasing the similarity between a sample and its transformed positive counterparts, and some also treat other samples as negative examples. Notable techniques in this category include memory bank \cite{chen2021exploring} and in-batch \cite{ye2019unsupervised} sampling. On the other hand, some methods use positive pairs with Siamese networks \cite{grill2020bootstrap} or add a relationship module \cite{patacchiola2020self} to perform the self-supervised task.

Transformation-based self-supervision is another popular approach, which involves generating new classes with data augmentation, predicting relative positions of patches, solving puzzles, or predicting rotations. One unique method in this category is LoRot \cite{moon2022tailoring}, designed for a different objective: to aid supervised learning.

Recently, there have been attempts to transfer the benefits of self-supervised learning to supervised learning. SupCLR \cite{khosla2020supervised} adapted the relation-based self-supervised framework to utilize labeled data since class labels clearly define both positive and negative examples. Additionally, self-label augmentation (SLA) \cite{gao2022decoupled} expanded the label space by combining the supervised class labels with data transformation labels, as using auxiliary pretext tasks can decrease performance. On the other hand, LoRot is a self-supervised method that can be directly applied to improve supervised learning.

\textbf{Mixture of Experts:} Sparsely-gated MoE \cite{shazeer2017} is the first model to show major improvements in model capacity, training time, or model quality upon activation. Switch Transformer \cite{fedus2021switch} simplifies activation by selecting only the best expert for each token using softmax in the hidden state and exhibits better scalability than previous works. All previous work required ancillary losses to encourage balance. This provision for loss must be carefully considered not to overwhelm the original loss. However, the auxiliary loss does not guarantee balance, and a hard power factor must be applied. Therefore, many tokens may not be affected by the MoE layer. Hard MoE \cite{gross2017hard} with a single decoding layer can be trained effectively in large-scale hashtag prediction tasks. Base Layers \cite{pmlr-v139-lewis21a} construct a linear assignment to maximize the token-expert relationship while ensuring that each expert receives an equal number of tokens. Hash classes design hashing techniques on input tokens. Unlike previous works, our method is a learned one that enables heterogeneous MoE and effectively improves downstream fine-tuning performance. We use MoE as a way to control the effect of each self-supervised transformation on the loss function. Thus, we can ensure that the model focuses on the most relevant augmentations for classification. Our method is similar to  \cite{yudistira2017gated} where the gates are attached to the last classification layer to fuse the two-stream CNN backbone on video classification. This approach differs from previous self-supervised methods, which typically rely on a single task or set of tasks and do not have a way to adjust the importance of different augmentations.

\section{Methodology}
\label{sec:pagestyle}

\begin{figure}
    \centering
    \includegraphics[width=1.0\linewidth]{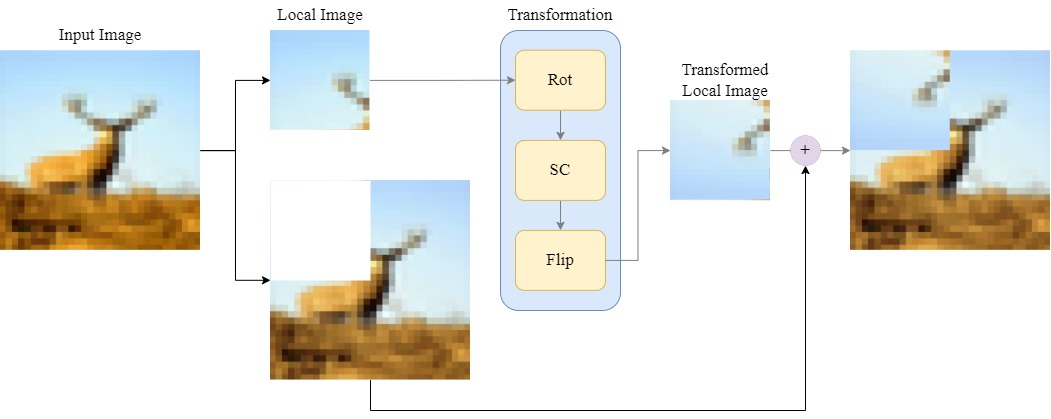}
    \caption{Diagram of Transformation}
    \label{fig:diagram_transformation}
\end{figure}
We first discussed each transformation we used as data augmentation in Sec. \ref{ssec:transformation}. Then, we used LDAM-DRW to train the imbalanced dataset in Sec. \ref{ssec:ldam-drw}. Finally, we proposed Gated Self-Supervised Learning based on a Mixture of Expert approaches, allowing the model to focus on relevant transformations.

\subsection{Transformation}
\label{ssec:transformation}
The first instances of the usefulness of data augmentation were shown using simple modifications like flipping images horizontally, changing the color space, and applying random croppings (Figure \ref{fig:diagram_transformation}). These techniques address issues related to image recognition tasks, such as insensitivity to spatial changes. This section will cover various augmentations based on geometric transformations and other image-processing methods. The augmentations discussed are notable for their simplicity of implementation. Understanding these basic transformations will serve as a foundation for exploring more advanced data augmentation techniques.

Our proposed method uses a combination of data augmentation, LDAM-DRW, and a Mixture of Expert (MoE) approaches to improve image classification performance on imbalanced datasets.
We utilized the different transformations as data augmentations, including rotation (using Lorot-E \cite{moon2022tailoring}), flip, and shuffle channels. These augmentations are chosen as they are easy to implement yet are effective in encoding invariances and challenges present in image recognition tasks. Then, LDAM-DRW is used to train imbalanced datasets, which improves the performance and accuracy of the model by balancing the distribution of the classes. Finally, the MoE is attached to gating every self-supervised task. This allows the model to learn the importance of each transformation and dynamically adjust the weighting of each one for classification. Additionally, the gating network is built with a fully-connected layer with softmax activation to output a class for each transformation.

\subsubsection{Rotation}
\label{sssec:rotation}
We use Lorot-E\cite{moon2022tailoring} as the transformation for the rotation transformation. The image will be divided into four quadrants (2x2 grid), then randomly selected and rotated from {0, 90, 180, 270}. This transformation will create 16 classes. The rotation degree parameter heavily determines the degree of rotation augmentations.
\subsubsection{Flip}
\label{sssec:flip}
Flipping is one of the easiest to implement and has proven useful on datasets. This task will randomly flip the selected quadrant of the image along the x-axis. The number of classes resulting from this transformation is two classes.
\subsubsection{Shuffle Channel}
\label{sssec:sc}
This transformation is used to shuffle the arrangement of the RGB channel of the selected quadrant of the image. The number of classes resulting from this transformation is six classes (3P3 permutation).

    \begin{figure}
        \centering
        \includegraphics[width=1.0\linewidth]{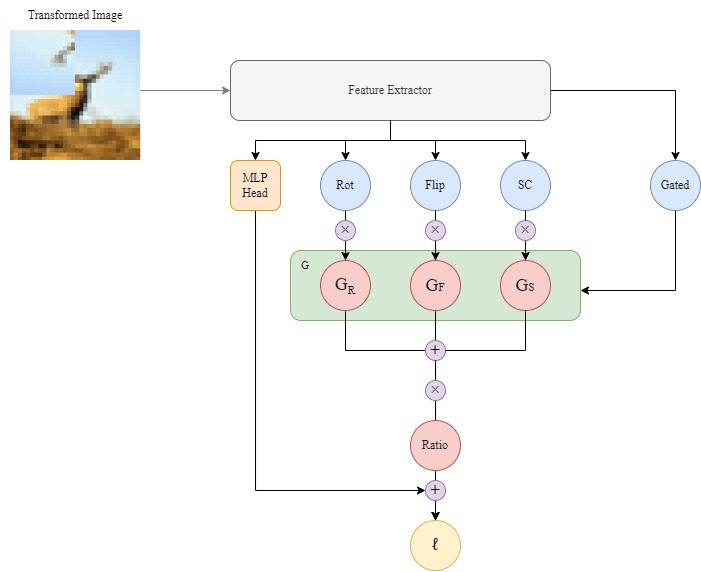}
        \caption{Diagram of Mixture of Expert}
        \label{fig:diagram_paper}
    \end{figure}

\subsection{LDAM-DRW}
\label{ssec:ldam-drw}

\begin{table*}\centering
\caption{Imbalanced classification accuracy (\%) in CIFAR-10/100}
\begin{tabular}{l|lll|lll}
\hline
\label{tab:results}
Imbalance Ratio & 0.01  & 0.02  & 0.05  & 0.01  & 0.02  & 0.05 \\ \hline
LDAM-DRW & 77.03 & 80.94 & 85.46 & 42.04 & 46.15 & 53.25 \\
+SSP & 77.83 & 82.13 & - & 43.43 & 47.11 & - \\
+SLA-SD & 80.24 & - & - & 45.53 & - & - \\
+LoRot-E & 81.82 & 84.41 & 86.67 & 46.48 & 50.05 & 54.66 \\ \hline
+MoE(LoRot-E+Flip) & 81.65 & 83.93 & 86.64 & 46.48 & 49.96 & 54.63  \\
+MoE(LoRot-E+ShuffleChannel) & \textbf{81.91} & 84.33 & \textbf{86.91} & \textbf{46.53} & 49.95 & \textbf{54.85} \\
+MoE(LoRot-E+Flip+ShuffleChannel) & 81.67 & \textbf{84.65} & 86.35 & \textbf{47.37} & \textbf{50.57} & \textbf{54.75} \\ \hline
\end{tabular}
\end{table*}

LDAM-DRW (Label-Distribution-Aware Margin - Deferred Re-Weighting) is a combination of two techniques designed to improve the performance of machine learning models in situations where the training data is heavily imbalanced among different classes, and the evaluation criteria require good generalization to the less common classes \cite{Cao2019}. In this experiment, we use the LDAM-DRW method with Gated Self-Supervision Method, which we propose to improve the performance and accuracy of the model.

\subsection{Mixture of Experts}

Mixture-of-Experts (MoE) is a type of deep learning architecture that combines multiple models, referred to as experts, to divide a complex task into simpler sub-problems that can each be addressed by an individual expert \cite{Chen2022}. In this work, we use MoE to gating every self-supervision task, which is used to learn the importance of each transformation used in self-supervised learning as illustrated in Figure \ref{fig:diagram_paper}. Each transformation has its linear head to output the class of the transformation. The weight gate of each transformation is also learned using an MoE, which allows the model to adjust the importance of each transformation for classification dynamically. We use a fully-connected layer with softmax activation function for the gating network as in eq. \ref{eq:softmax}:
\begin{equation}
    \label{eq:softmax}
    G = softmax(W^{T}X + b) 
\end{equation}
Where $G$, $X$, $W$, and $B$ refer to the gate, baseline output, weight gating network, and bias gating network, respectively. Specifically, we gates every loss from each self-supervision task $L$ and sum all the gating loss. Then, we combine the loss of the classifier (supervised) $L_{C}$ and the loss of the self-supervision as follows in eq. \ref{eq:gating}:
\begin{equation}
    \label{eq:gating}
    L_{tot} = L_{C}+ \lambda \sum_{n=1}^{t}G_{n}^{T}L_{n}
\end{equation}
where $t$ is the number of the self-supervision task and $\lambda$ is SSL ratio.

\section{Experimental Setup}
\label{sec:experimentalsetup}
In the imbalanced task experiment, we tested the gated self-supervised learning method using the Google Colab Pro environment. The GPU that we use for this experiment is Nvidia T4 GPU. For a fair comparison, we use the same backbone and baseline as previous research to train the model for the experiment. We use the Resnet-32 architecture for the backbone of the network and LDAM-DRW \cite{Cao2019} as the baseline and follows the baseline settings. We set the batch size to 128 and the epochs for training the model to 300. For the learning rate, we set the initial value to 0.1, which is dropped by a factor of 0.01 at the 160-$th$ epoch and 180-$th$ epoch. The optimizer we use in this experiment is Stochastic Gradient Descent with the momentum of 0,9 and weight decay $2\times10^{-4}$. For the SSL ratio, we set all of the experiments to 0.1.

We also train the model using our proposed method in the Tiny-Imagenet dataset \cite{le2015tiny} using a single GPU Nvidia Quadro RTX 8000. For a fair comparison, we use the same backbone and setups in each method that will be tested. To train the model, we set the batch size to 256 and the epochs for training the model to 300. Resnet18 will be used as the backbone, with Stochastic Gradient Descent as the optimizer. We set the momentum to 0,9 and weight decay to $2\times10^{-4}$. For the learning rate, we set the initial value to 0.1, decaying by a factor of 0.1 every 75 epochs. Finally, for the SSL ratio, we set all of the experiments to 0.1.

\begin{table}[]
\caption{Additional Experiments on Tiny-Imagenet}
\begin{tabular}{l|c}
\hline
\label{tab:results_tiny-imagenet}
                            & Val accuracy  \\ \hline
ResNet 18                         & 46.68          \\
+LoRot-E                          & 48.65          \\ \hline
+LoRot-E+Flip+ShuffleChannel      & 48.86          \\ \hline
+Moe(LoRot-E+Flip)                & 47.86          \\
+MoE(LoRot-E+ShuffleChannel)    & 48.52          \\
+MoE(LoRot-E+Flip+ShuffleChannel) & \textbf{48.99} \\ \hline
\end{tabular}
\end{table}

\section{Experiments}
\label{sec:experiments}

For the experiment, we use LDAM-DRW \cite{Cao2019} to do the Imbalanced Classification and Cifar dataset to design the imbalanced scenario. We create three combinations of gated self-supervised learning for the proposed method, such as Lorot-E+flip, Lorot-E+ShuffleChannel, and Lorot-E+FLip+ShuffleChannel. Moreover, we also vary the imbalance scenario of Cifar 10 and Cifar 100 datasets. We apply the imbalance ratio to 0.01, 0.02, and 0.05. To compare our proposed method with other self-supervision techniques, we report the result of LDAM-DRW\cite{Cao2019}, SSP\cite{yang2020rethinking}, SLA+SD \cite{lee2020self} and, Lorot-E\cite{moon2022tailoring}.  

The results of the imbalanced classification are shown in Table \ref{tab:results}. As we can see, all of the combination tasks for MoE have a clear complementary effect and improve the accuracy of the LDAM-DRW, SSP, and SLA+SD method, with a gain of up to +4.87\%. However, several combinations of MoE improve the Lorot-E method, such as Lorot-E + flip and Lorot-E + flip + shuffle channel. However, the MoE combination of Lorot+Flip can not improve the Lorot-E model in two imbalanced datasets, such as Cifar 10 and Cifar 100, thus reducing the model's accuracy. Meanwhile, the other combinations of Lorot-E + ShuffleChannel, and Lorot-E+flip+ShuffleChannel improve the accuracy of the Lorot-E model in the imbalanced Cifar 10 ad Cifar 100. Specifically, the Lorot-E+flip+Shuffle-Channel combination successfully improves the model's accuracy in all imbalanced scenarios in Cifar 100 dataset. Therefore, Lorot-E+flip+ShuffleChannel is the best combination to improve the model's accuracy in the imbalanced task.

In order to further evaluate the effectiveness of our proposed method, we tested our method on the Tiny-Imagenet dataset. We found that it consistently outperformed other methods, as shown in Table \ref{tab:results_tiny-imagenet}. This was particularly impressive given the large number of classes in the Tiny-Imagenet dataset, as our method improved even on this challenging dataset. Overall, our results demonstrate the versatility and effectiveness of our method, as it was able to achieve better performance on all datasets tested, so our method has the potential to be widely applicable and useful for a variety of image classification tasks.

This work considers each of these transformations as a non-linear one. Therefore, the gated self-supervised learning method can tackle the problem of determining which transformation is important for the model to learn. However, to make further improvements, we need to find and select the combination self-supervision task of MoE that will best improve the model's accuracy.





\section{Conclusion}
\label{sec:majhead}


Self-supervised learning has gained increasing attention in recent years as a way to train deep learning models using large amounts of unlabeled data. The main idea behind self-supervised learning is to use the inherent structure in the data to create a supervised learning problem that can be used to train a model. In this work, we proposed a self-supervised learning method that adds additional transformations, such as flip and shuffle channels, to the past method, rotation transformation. Our results show that adding these additional transformations helps to increase the accuracy of the model, especially when using larger datasets such as CIFAR 100. Furthermore, compared to other methods, our proposed method gave better accuracy in all experiments when using CIFAR 100. On CIFAR 10, our method also gave better accuracy in certain experiments. Finally, on Tiny-Imagenet, we achieve better results using a self-supervised learning method on top of the model.

Overall, our results suggest that adding more transformations to self-supervised learning, in combination with the gating method, can help improve the model's performance. This highlights the potential of self-supervised learning as a powerful tool for training deep learning models in various settings.


\vfill\pagebreak


\bibliographystyle{IEEEbib}
\bibliography{export}

\end{document}